\documentclass[article,12pt]{elsarticle}

\usepackage{amssymb}
\usepackage{graphicx}
\usepackage{amsmath}
\usepackage{url}
\usepackage{algpseudocode}
\usepackage{algorithm}
\usepackage{booktabs}
\usepackage{pdflscape}
\usepackage{afterpage}
\usepackage{rotating}
\usepackage{color}

\newtheorem{thm}{Theorem}
\newtheorem{lem}[thm]{Lemma}
\newdefinition{rmk}{Remark}
\newproof{pf}{Proof}

\newdefinition{definition}{Definition}
\newdefinition{example}{Example}

\begin{document}

\begin{frontmatter}  

\title{Solving the Minimum Common String Partition Problem with the Help of Ants}


%
%
\author[buet] {S. M. Ferdous }
\author[buet] {M. Sohel Rahman }
%

\address[buet]{A$\ell$EDA Group, Department of CSE, BUET, Dhaka-1000, Bangladesh}

%
%


\begin{abstract}
In this paper, we consider the problem of finding a minimum common partition of two strings. The problem has its application in genome comparison. As it is an NP-hard, discrete combinatorial optimization problem, we employ a metaheuristic technique, namely, MAX-MIN ant system to solve this problem. To achieve better efficiency we first map the problem instance into a special kind of graph. Subsequently, we employ a MAX-MIN ant system to achieve high quality solutions for the problem. Experimental results show the superiority of our algorithm in comparison with the state of art algorithm in the literature. The improvement achieved is also justified by standard statistical test.
\end{abstract}

\end{frontmatter}

\section{Introduction}

String comparison is one of the important problems in Computer Science with diverse applications in different areas including Genome Sequencing, text processing and compressions. In this paper, we address the problem of finding a minimum common partition (MCSP) of two strings. MCSP is closely related to genome arrangement which is an important topic in computational biology. Given two DNA sequences, the MCSP asks for the least-sized set of the common building blocks of the sequences.  

In the MCSP problem, we are given two \emph{related} strings $(X,Y)$. Two strings are related if every letter appears the same number of times in each of them. Clearly, two strings have a common partition if and only if they are related. So, the length of the two strings are also the same (say, $n$). Our goal is to partition each string into $c$ segments called $blocks$, so that the $blocks$ in the partition of $X$ and that of $Y$ constitute the same multiset of substrings. Cardinality of the partition set, i.e., $c$ is to be minimized. A partition of a string $X$ is a sequence $P = (B_1,B_2, \cdot\cdot\cdot ,B_m)$ of strings whose
concatenation is equal to $X$, that is $B_1B_2 \cdot\cdot\cdot B_m = X$. The strings $B_i$ are called
the blocks of $P$. Given a partition $P$ of a string $X$ and a partition $Q$ of a string
$Y$, we say that the pair $\pi = <P,Q>$ is a common partition of $X$ and $Y$ if $Q$ is a
permutation of $P$. The minimum common string partition problem is to find a
common partition of $X$, $Y$ with the minimum number of blocks. For example, if $(X,Y)$ = \{``ababcab'',``abcabab''\}, then one of the minimum common partition sets is $\pi = $\{``ab'',``abc'',``ab''\} and the minimum common partition size is 3. The restricted
version of MCSP where each letter occurs at most $k$ times in each input string, is denoted by $k$-MCSP.

MCSP has its vast application rooted in Comparative Genomics. Given two DNA strings, MCSP answers the possibilities of re-arrangement of one DNA string to another \cite{peter}. MCSP is also important in ortholog assignment. In\cite{chen}, the authors present a new approach to ortholog assignment that takes into account both sequence similarity and evolutionary events at a genomic level. In that approach, first, the problem is formulated as that of computing the signed reversal distance with duplicates between the two genomes of interest. Then, the problem is decomposed into two optimization problems, namely minimum common partition and maximum cycle decomposition problem. Thus MCSP plays an integral part in computing ortholog assignment of genes.

\subsection{Our Contribution}
In this paper, we consider metaheuristic approaches to solve the problem. To the best of our knowledge, there exists no attempt to solve the problem with metaheuristic approaches. Only theoretical works are present in literature. Particularly we are interested in nature inspired algorithms. As the problem is  discrete combinatorial optimization problem, the natural choice is Ant Colony Optimization (ACO). Before applying ACO, it is necessary to map the problem into a graph. We have developed this mapping. In this paper, we implement a variant of ACO algorithm namely MAX-MIN Ant System (MMAS) to solve the MCSP problem. We conduct experiments on both random and real data to compare our algorithm with the state of the art algorithm in the literature and achieve excellent results. Notably, a preliminary version of the paper appeared at~\cite{FerdousR13}.

\section{Literature Review}
MCSP is essentially the breakpoint distance problem \cite{chromosome} between two permutations which is to count the number of ordered pairs of symbols that are adjacent in the first string but not in the other; this problem is obviously solvable in polynomial time \cite{goldstein}. The 2-MCSP is proved to be NP-hard and moreover  APX-hard in \cite{goldstein}. The authors in \cite{goldstein} also presented several approximation algorithms. Chen et al. \cite{chen} studied the problem, Signed Reversal Distance with Duplicates (SRDD), which is a generalization of MCSP. They gave a 1.5-approximation algorithm for 2-MCSP. In \cite{peter}, the author analyzed the fixed-parameter tractability of MCSP considering different parametrs.
In \cite{jiang}, the authors investigated $k$-MCSP along with two other variants: $MCSP^c$, where the alphabet size is at most $c$; and $x$-balanced MCSP, which requires that the length of the blocks must be within the range $(n/d - x, n/d + x)$,
where $d$ is the number of blocks in the optimal common partition and $x$ is a constant integer. They showed that
$MCSP^c$ is NP-hard when $c \geq 2$. As for $k$-MCSP, they presented an FPT
algorithm which runs in $O^*((d!)^{2k})$ time.

Chrobak et al. \cite{chrobak} analyzed a natural greedy heuristic for MCSP: iteratively, at each step, it extracts a longest common substring from the input strings. They showed that for 2-MCSP, the approximation ratio (for the greedy heuristic) is exactly 3.
They also proved that for 4-MCSP the ratio would be $\log n$ and for the general MCSP, between $\Omega(n^{0.43})$ and $O(n^{0.67})$.

Ant colony optimization (ACO) \cite{jour_dorigo, jour1_dorigo, book_dorigo} was introduced by
M. Dorigo and colleagues as a novel nature-inspired metaheuristic for the solution of hard combinatorial optimization (CO) problems. The inspiring source of ACO is the pheromone trail laying and following behavior of real ants which use
pheromones as a communication medium. In analogy to the biological example,
ACO is based on the indirect communication of a colony of simple agents, called
(artificial) ants, mediated by (artificial) pheromone trails. The pheromone trails
in ACO serve as a distributed, numerical information which the ants use to probabilistically
construct solutions to the problem being solved and which the ants
adapt during the algorithm's execution to reflect their search experience.

Different ACO algorithms have been proposed in the literature. The original algorithm is known as the Ant System(AS) \cite{pos_dorigo,dis_dorigo,jour3_dorigo}. The other variants are, Elitist AS \cite{dis_dorigo,jour3_dorigo}, ANT-Q \cite{antq}, Ant Colony System (ACS) \cite{jour1_dorigo}, MAX-MIN AS \cite{mmas1,mmas2,jour_Utzle} etc.

Recently growing interest has been noticed towards ACO in the scientific community. There are now available several successful implementations of the ACO metaheuristic applied to a number of different discrete combinatorial optimization problems.
In \cite{jour_dorigo} the authors distinguished among two classes of applications of ACO: those to static combinatorial optimization problems, and those to the dynamic
ones. When the problem is defined and does not change while the problem is being solved is termed as static combinatorial optimization problems. The authors list some static combinatorial optimization
problems those are successfully solved by different variants of ACO. Some of the problems are, travelling salesperson, Quadratic Assignment,
job-shop scheduling, vehicle routing, sequential ordering, graph coloring etc. Dynamic problems
are defined as a function of some quantities whose values are set by the dynamics of
an underlying system. The problem changes therefore at run time and the optimization algorithm must be capable of adapting online to the changing environment. The authors listed connection-oriented network routing and
connectionless network routing as the examples of dynamic problems those are successfully solved by ACO.

In 2010 a non-exhaustive list of applications of ACO algorithms grouped by problem types is presented in \cite{survey_dorigo_2010}. The authors categorized the problems into different types
namely routing, assignment, scheduling, subset machine learning and bioinformatics. In each type they listed the problems those are successfully solved by some variants of ACO.

There are not too many string related problems solved by ACO in the literature. In \cite{blum_seq}, the authors addressed the reconstruction of DNA sequences from DNA fragments by ACO. Several ACO algorithms have been proposed for the longest common subsequence (LCS) problem in \cite{lcs_aco_shyu,lcs_aco_christ}. Recently minimum string cover problem is solved by ACO in \cite{mscp_aco}. Finally, we note that a preliminary version of this work was presented at \cite{confVersion}.
\section{Preliminaries}

In this section, we present some definitions and notations that are used throughout the paper.
\begin{definition} \emph{{Related string}:}
Two strings $ (X,Y) $, each of length $n$, over an alphabet $\sum $ are called \emph{related} if every letter appears the same number of times in each of them.
\end{definition}

\begin{example}
    $X$ = ``abacbd'' and $Y$ = ``acbbad'', then they are \emph{related}. But if $X_1$ = ``aeacbd'' and $Y$ = ``acbbad'', they are not \emph{related}
\end{example}

\begin{definition}{\emph{Block}:}
 A block $B=([id,i,j])$, $0\leq i\leq j < n$, of a string $S$ is a data structure having three fields: $id$ is an identifier of $S$ and the starting and ending positions of the block in $S$ are represented by $i$ and $j$, respectively. Naturally, the \emph{length} of a block $[id,i,j]$ is $(j-i+1)$. We use $substring([id,i,j])$ to denote the substring of $S$ induced by the block $[id,i,j]$. Throughout the report we will use 0 and 1 as the identifiers of $X$(i.e., $id(X)$) and $Y$(i.e., $id(Y)$) respectively. We use $[]$ to denote an empty block.
 \end{definition}

\begin{example}
If we have two strings $(X,Y)$ = \{``abcdab'',``bcdaba''\}, then $[0,0,1]$ and $[0,4,5]$ both represent the substring ``ab'' of $X$. In other words, $substring([0,0,1]) = substring([0,4,5]) =$ ``ab''.
\end{example}

Two blocks can be intersected or unioned. The intersection of two blocks (with same ids) is a block that contains the common portion of the two.
\begin{definition}{\emph{Intersection of blocks}:}
Formally, the intersection operation of $B_1$=$[id,i,j]$ and $B_2$=$[id,i',j']$ is defined as follows:
\begin{equation}
B_1 \cap B_2 = \left\{
  \begin{array}{l l}
    []  & \quad \text{if $i' > j$ or $i > j'$}\\
    $[$id,i',j$]$ & \quad \text{if $i' \leq j$}\\
    $[$id,i,j'$]$ & \quad \text{else}
  \end{array} \right.
\end{equation}
\end{definition}

\begin{example}
    If, $B_1 = [0,1,5]$ and $B_2 = [0,3,6]$, then $B_1 \cap B_2 = [0,3,5]$. On the other hand, if  $B_1 = [0,1,5]$ and $B_2 = [0,6,8]$, then $B_1 \cap B_2 = []$
\end{example}
\begin{definition}{\emph{Union of blocks}:}
Union of two blocks (with same ids) is either another block or an ordered (based on the starting position) set of blocks. Without the loss of generality we suppose that,  $i \leq i'$ for $B_1$=$[id,i,j]$ and $B_2$=$[id,i',j']$. Then, formally the union operation of $B_1$ and $B_2$ is defined as follows:
\begin{equation}
B_1 \cup B_2 = \left\{
  \begin{array}{l l}
    $[$id,i,j$]$  & \quad \text{if $j' \leq j$}\\
    $[$id,i,j'$]$ & \quad \text{if $j' > j$ or $i' = j+1$}\\
    $\{$B_1,B_2$\}$ & \quad \text{else}
  \end{array} \right.
\end{equation}
\end{definition}

\begin{example}
    If, $B_1 = [0,1,5]$ and $B_2 = [0,3,6]$, then $B_1 \cup B_2 = [0,1,6]$. On the other hand, if  $B_1 = [0,1,5]$ and $B_2 = [0,6,8]$, then $B_1 \cup B_2 = \{[0,1,5],[0,6,8]\}$
\end{example}

The union rule with an ordered set of blocks, $B_{lst}$ and a block, $B'$ can be defined as follows. We have to find the position where $B'$ can be placed in $B_{lst}$, i.e., we have to find $B_k \in B_{lst}$ after which $B'$ can be placed. Then, we have to replace the ordered subset $\{B_k,B_{k+1}\}$ with $B_k \cup B' \cup B_{k+1}$.

\begin{example}
As an example, suppose we have three blocks, namely, $B_1 = [0,5,7]$,$B_2 = [0,11,12]$ and $B_3 = [0,8,10]$. Then $B_1 \cup B_2 = B_{lst}' = \{[0,5,7],[0,11,12]\}$. On the other hand, $B_{lst}' \cup B_3 = [0,5,12]$, which is basically identical to $B_1 \cup B_2 \cup B_3$.
\end{example}

Two blocks $B_1$ and $B_2$ (in the same string or in two different strings) match if $substring(B_1) = substring(B_2)$. If the two matched blocks are in two different strings then the matched substring is called a common substring of the two strings denoted by \emph{cstring($B_1,B_2$)}.

\begin{definition}{\emph{span}:}
Given a list of blocks with same id, the \emph{span} of a block,  $B = [id,i,j]$ in the list denoted by, $span(B)$ is the length of the block (also in the list) that contains $B$ and whose length is maximum over all such blocks in the list. Note that a block is assumed to contain itself. More formally, given a list of blocks, $list_{b}$, $span(B \in list_{b}) =\max\{ \ell~|~ \ell= length(B'), B \subseteq B', \forall B' \in list_{b}\}$.
\end{definition}

\begin{example}
 If $list_{b}=\{[0,0,0],[0,0,1],[0,0,2],[0,4,5]\}$ then $span([0,0,0]) = span([0,0,1]) = span([0,0,2]) = 3$ where as, $span([0,4,5]) = 2$. In other words, span of a block is the maximum length of the super string than contains the substring induced by the block.
\end{example}

\begin{definition}{\emph{Partition}:}
A \emph{partition}  of a string $X$ is a list of blocks all with $id(X)$ having the following two properties:

\begin{enumerate}[(a)]
\item Non Overlapping: The blocks must be be disjoint, i.e., no block should overlap with another block. So the intersection of any two blocks must be empty.
\item Cover: The blocks must cover the whole string.
\end{enumerate}

In other words, a \emph{partition} of a string $X$ is a sequence $P = (B_1,B_2, \ldots ,B_m)$ of strings whose
concatenation is equal to $X$, that is $B_1B_2 \ldots B_m = X$. where $B_i$'s are \emph{blocks}.
\end{definition}

\subsection{Basics of ACO}

In ACO,  a combinatorial optimization (CO) problem is solved by iterating the following two steps. At first, solutions are constructed using a parameterized probability distribution over the solution space which is called pheromone model. The second step is to modify the pheromone values using the solutions that were constructed in earlier iterations in a way that is deemed to bias the search towards the high quality solutions.

\subsection{Ant Based Solutions Construction}
 The basic ingredient of an ACO algorithm is a constructive heuristic that constructs solutions probabilistically. Sequences of solution components taken from a finite set of solution components $C = \{c_1,c_2,...c_n\}$ is assembled by a constructive heuristic. Starting with an empty partial solution $s^p = \emptyset$ a solution is constructed. Then at each construction step the current partial solution $s^p$ is extended by adding a feasible solution component from the solution space $C$. The definition of feasible solution component is problem specific. Typically a problem is mapped into a construction Graph $G_c = (C,E)$ whose vertices are the solution components $C$ and the set $E$ are the connections (i.e., edges). The process of constructing solutions can be regarded as a walk (or a path) on the construction graph.

\subsection{Heuristic Information}
In most ACO algorithms the transition probabilities, i.e., the probabilities for choosing the next solution component, are defined as follows:
\begin{equation}
    p(c_i|s^p) = \frac{{\tau_i}^\alpha \cdot \eta(c_i)^\beta}{\sum_{c_j \in N(s^p)}{\tau_j}^\alpha \cdot \eta(c_j)^\beta}, \forall c_i \in N(s^p)
\end{equation}
Here,
$c_i$ is a candidate component, $s^p$ is the partial solution. The current partial solution $s^p$ is extended
by adding a feasible solution component from the set of feasible neighbors $N(s^p) \subseteq C$. $\eta$ is a weight function that contains \emph{heuristic information} and $\alpha, \beta$ are positive parameters whose values determine the relation between the \emph{pheromone information} and the \emph{heuristic information}. The pheromones deployed by the ants are denoted by $\tau$.

\subsection{Pheromone Update}
The pheromone update consists of two parts. The first part is pheromone evaporation, which uniformly decreases all the pheromone values . From a practical point of view, pheromone evaporation prevents too rapid convergence of the algorithm toward a sub-optimal region. Thus it helps to avoid the local optimal solutions and favors the exploration of new areas in the search space. Then, one or more solutions from the current or from earlier iterations (the set is denoted by $S_{upd}$)are used to increase the values of pheromone trail parameters on solution components that are part of these solutions:
\begin{equation}
    \label{pheromone}
    \tau_i \leftarrow (1-\varepsilon) \times \tau_i + \varepsilon \times \sum_{s \in S_{upd}|c_i \in s}F(s) , i=1,2,...,n
\end{equation}

Let $W(.)$ is the cost function. Here, $S_{upd}$ is the set of local best or global best solution, $\varepsilon \in (0,1]$ is a parameter called the \emph{evaporation rate}, and $F: G \rightarrow \mathbb{R^+}$ is a function such that $W(s)<W(\acute{s})\Rightarrow F(s) \geq F(\acute{s}), s \neq \acute{s}, \forall s \in G$. The function $F(.)$ is commonly called the \emph{Fitness Function}.

In general, different versions of ACO algorithms differ in the way they update the pheromone values. This also holds for the two currently best-performing ACO variants in practice, namely, the Ant Colony System (ACS) \cite{jour1_dorigo} and the MAX-MIN Ant System (MMAS) \cite{jour_Utzle}. Since in our algorithm we hybridize ACS with MMAS, below we give a brief description of MMAS.

\subsection{MAX-MIN Ant System (MMAS)}
MMAS algorithms are characterized as follows. First, the pheromone values are limited to an interval $[\tau_{MIN},\tau_{MAX}]$ with $0 < \tau_{MIN} < \tau_{MAX}$. Pheromone trails are initialized to $\tau_{max}$ to favor the diversification during the early iterations so that premature convergence is prevented. Explicit limits on the pheromone values ensure that the chance of finding a global optimum never becomes zero. Second, in case the algorithm detects that the search is too much confined to a certain area in the search space, a restart is performed. This is done by initializing all the pheromone values again. Third, the pheromone update is always performed with either the iteration-best solution, the restart-best solution (i.e., the best solution found since the last restart was performed), or the best-so-far solution.


\section{Our Approach: MAX-MIN Ant System on the Common Substring Graph}
\subsection{Formulation of \emph{Common Substring Graph}}
We define a common substring graph, $G_{cs}(V,E,id(X))$ of a string $X$ with respect to $Y$ as follows. Here $V$ is the vertex set of the graph and $E$ is the edge set. Vertices are the positions of string $X$, i.e., for each $v \in V$, $v \in [0,|X|-1]$. Two vertices $v_i \leq v_j$ are connected with and edge, i.e, $(v_i, v_j) \in E$, if the substring induced by the block $[id(X),v_i,v_j]$ matches some substring of $Y$. More formally,  we have:

$$
    (v_i,v_j) \in E \Leftrightarrow cstring([id(X),v_i,v_j],B')\ is\ not\ empty\ \   \exists{B'\in Y}
$$

In other words, each edge in the edge set corresponds to a \emph{block} satisfying the above condition. For convenience, we will denote the edges as \emph{edge blocks} and use the list of edge blocks (instead of edges) to define the edgeset $E$. Notably, each \emph{edge block} on the edge set of $G_{cs}(V,E,id(X))$ of string $(X,Y)$ may match with more than one blocks of $Y$. For each \emph{edge block} $B$ a list is maintained containing all the matched blocks of string $Y$ to that \emph{edge block}. This list is called the $matchList(B)$.

\begin{figure}
    \begin{center}
  \includegraphics[width=0.8\textwidth]{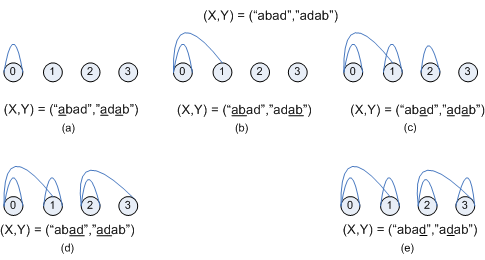}
  \caption{Construction of $G_{cs}(V,E,id(X))$ of $(X,Y)$. (a) Vertex 0 is connected with itself because ``a'' is common string of $X$ and $Y$ (b) An edge between vertices 0 and 1 as ``ab'' is a common string of $X$ and $Y$. (c) vertex 1 is connected with itself (d) vertex 1 and 2 are connected with (e) Vertex 3 is connected with itself.}
  \label{graph}
  \end{center}
\end{figure}

For example, suppose $(X,Y)$ = \{``abad'',``adab''\}. Now consider the corresponding common substring graph, $G_{cs}(V,E,id(X))$. Then, we have $V=\{0,1,2,3\}$ and $E = \{[0,0,0],[0,0,1],[0,1,1],[0,2,2],[0,2,3]\}$. The construction steps are shown in figure \ref{graph}.

To find a common partition of two strings ($X,Y$) we first construct the common substring graph of $(X,Y)$. Then from a vertex $v_i$ on the graph we take an edge block $[id(X),v_i,v_j]$. Suppose $M_i$ is the $matchList$ of this block. We take a block $B'_i$ from $M_i$. Then we advance to the next vertex that is ($v_j+1)\ MOD\ |X|$ and choose another corresponding edge block as before. We continue this until we come back to the starting vertex. Let $partitionList$ and $mappedList$ are two lists, each of length $c$, containing the traversed edge blocks and the corresponding matched blocks. Now we have the following lemma.

\begin{lem}
$partitionList$ is a common partition of length $c$ iff,
\begin{equation}
\label{eq1}
    B_i \cap B_j = []\ \forall{B_i,B_j \in mappedList}, \ i \neq j\
\end{equation}
and
\begin{equation}
\label{eq2}
    B_1 \cup B_2 \cup \cdot\cdot\cdot \cup B_c = [id(Y),0,|Y|-1]
\end{equation}
\end{lem}

\begin{pf}
    By construction, $partitionList$ is a \emph{partition} of $X$. We need to prove that $mappedList$ is a partition of $Y$ and with the one to one correspondence between $partitionList$ and $mappedList$ it is obvious that $partitionList$ would be the common partition of $(X,Y)$. Equation \ref{eq1} asserts the non overlapping property of $mappedList$ and Equation \ref{eq2} assures the cover property. So, $mappedList$ will be a partition of $Y$ if Equation \ref{eq1} and \ref{eq2} are satisfied.

    On the other hand let $partitionList$ along with $mappedList$ is a common partition of $(X,Y)$. According to construction, $partitionList$ satisfies the two properties of a partition. Let, $mappedList$ is a partition of $Y$. We assume $mappedList$ does not follow the Equation \ref{eq1} or \ref{eq2}. So, there might be overlapping between the blocks or the blocks do not cover the string $Y$, a contradiction. This completes the proof.

\end{pf}

\subsection{Heuristics}
Heuristics ($\eta$) contain the problem specific information. We propose two different (types of) heuristics for MCSP. Firstly, we propose a static heuristic that does not change during the runs of algorithm. The other heuristic we propose is dynamic in the sense that it changes between the runs.
\subsubsection{The Static Heuristic for MCSP}
We employ an intuitive idea. It is obvious that the larger is the size of the blocks the smaller is the partition set. To capture this phenomenon, we assign on each edge of the common substring graph a numerical value that is proportional to the length of the substring corresponding to the edge block. Formally, the static heuristic ($\eta_s$) of an edge block $[id,i,j]$ is defined as follows:
\begin{equation}
    \eta_s([id,i,j]) \propto length([id,i,j])
\end{equation}

\subsubsection{The Dynamic Heuristic for MCSP}\label{sec:dynamicH}
We observe that the static heuristic can sometimes lead us to very bad solutions. For example if $(X,Y)$ = \{``bceabcd'',``abcdbec''\}
then according to the static heuristic much higher value will be assigned to \emph{edge block} $[0,0,1]$ than to $[0,0,0]$. But if we take $[0,0,1]$, we must match it to the block $[1,1,2]$ and we further miss the opportunity to take $[0,3,6]$ later. The resultant partition will be \{``bc'',``e'',``a'',``b'',``c'',``d''\} but if we would take $[0,0,0]$ at the first step, then one of the resultant partitions would be \{``b'',``c'',``e'',``abcd''\}. To overcome this shortcoming of the static heuristic we define a dynamic heuristic as follows. The dynamic heuristic ($\eta_{d}$) of an edge block ($B = [id,i,j]$) is inversely proportional to the difference between the length of the block and the minimum span of its corresponding blocks in its $matchList$. More formally, $\eta_d(B)$ is defined as follows:

\begin{equation}
    \eta_{d}(B) \propto \frac{1}{|length(B) - minSpan(B)|+1},
\end{equation}
where
\begin{equation}
\label{span}
minSpan(B) = \min\{span(B')~|~ B' \in matchList(B) \}
\end{equation}

In the example, $minSpan([0,0,0])$ is 1 as follows: $matchList([0,0,0])=\{ [1,1,1],[1,4,4]\}$. $span([1,1,1])=4$ and $span([1,4,4] = 1)$. On the other hand, $minSpan([0,0,1])$ is 4. So, according to the dynamic heuristic much higher numeral will be assigned to block $[0,0,0]$ rather than to block $[0,0,1]$.

We define the total heuristic ($\eta$) to the linear combination of the static heuristic ($\eta_s$) and the dynamic heuristic ($\eta_d$). Formally, the total heuristic of an edge block B is,

\begin{equation}
    \eta(B) = a \cdot \eta_s(B) + b \cdot \eta_d(B)
\end{equation}

where $a$, $b$ are any real valued constant. The algorithms of static and dynamic heuristics are shown in Algorithm (\ref{dynamic} - \ref{static})

\begin{algorithm}
\caption{addDynamicHeuristic($G_{cs}$)}
\label{dynamic}
\begin{algorithmic}
    \State E $\gets$ edge blocks of E
    \ForAll{Block B in E}
        \State minspan $\gets$ find minimum free span of B by Equation \ref{span}
        \State dynamicHeuristic(E) = $\frac{1}{(length(E) - minspan + 1)}$
    \EndFor
\end{algorithmic}
\end{algorithm}

\begin{algorithm}
\caption{addStaticHeuristic($G_{cs}$)}
\label{static}
\begin{algorithmic}
    \State E $\gets$ edge blocks of $G_{cs}$
    \State max $\gets$ maximum length edgeblock of $G_{cs}$
    \ForAll{Block B in E}
        \State staticHeuristic(B) = length(B)/max
    \EndFor
\end{algorithmic}
\end{algorithm}

\begin{algorithm}
\caption{addHeuristic($G_{cs}$,a,b)}
\label{heu}
\begin{algorithmic}
    \State E $\gets$ edge blocks of $G_{cs}$
    \State  addStaticHeuristic($G_{cs}$)
    \State  addDynamicHeuristic($G_{cs}$)
    \ForAll{Block B in E}
        \State heuristic(B) $\gets$ a $\cdot$ staticHeuristic(B) + b $\cdot$ dynamicHeuristic(B)
    \EndFor
\end{algorithmic}
\end{algorithm}

\subsection{Initialization and Configuration}
Given two strings $(X,Y)$, we first construct the common substring graph $G_{cs} = (V,E,id(X))$. We use the following notations. \emph{Local best solution} ($L_{LB}$) is the best solution found in each iteration. \emph{Global best solution} ($L_{GB}$) is the best solution found so far among all iterations. The pheromone of the edge block is bounded between $\tau_{max}$ and $\tau_{min}$. Like \cite{jour_Utzle}, we use the following values for $\tau_{max}$ and $\tau_{min}$: $\tau_{max} = \frac{1}{\varepsilon \cdot cost(L_{GB})}$, and $\tau_{min} = \frac{ \tau_{max}(1-\sqrt[n]{p_{best}})}{(avg-1)\sqrt[n]{p_{best}}}$. Here, $avg$ is the average number of choices an ant has in the construction phase; $n$ is the length of the string; $p_{best}$ is the probability of finding the best solution when the system converges and $\varepsilon$ is the evaporation rate. Initially, the pheromone values of all edge blocks (substring) are initialized to $initPheromone$ which is a large value to favor the exploration at the first iteration \cite{jour_Utzle}. The steps of the initialization is shown in Algorithm \ref{init}

\begin{algorithm}
\caption{initialize($G_{cs}$)}
\label{init}
\begin{algorithmic}
    \State initialize $L_{LB}$
    \State initialize $L_{GB}$
    \State set Parameters
    \State E $\gets$ edge blocks of $G_{cs}$

    \ForAll{ Block B in E}
        \State pheromone(B) $\gets$ $initPheromone$
    \EndFor
\end{algorithmic}
\end{algorithm}

\subsection{Construction of a Solution}

Let, $nAnts$ denotes the total number of ants in the colony. Each ant is deployed randomly to a vertex $v_s$ of $G_{cs}$. A solution for an ant starting at a vertex $v_s$ is constructed by the following steps:

    \textbf{\emph{step 1}}: Let $v_i = v_s$. Choose an \emph{available} edge block starting from $v_i$ by the discrete probability distribution defined below. An edge block is available if its $MatchList$ is not empty and inclusion of it to the $partitionList$ and $mappedList$ obeys Equation \ref{pdf}. The probability for choosing edge block $[0,v_i,v_j]$ is:

    \begin{equation}
    \label{pdf}
        p([0,v_i,v_j]) = \frac{\tau([0,v_i,v_j])^\alpha \cdot \eta([0,v_i,v_j])^\beta}{\sum_\ell {\tau([0,v_i,v_\ell])^\alpha \cdot \eta([0,v_i,v_\ell])^\beta}\  },
        \forall \ell \ such \ that [0,v_i,v_l]\ is \ an\ available\ block.
    \end{equation}

    \textbf{\emph{step 2}}: Suppose, $[0,v_i,v_k]$ is chosen according to Equation \ref{pdf} above. We choose a match block $B_m$ from the $matchList$ of $[0,v_i,v_k]$ and delete $B_m$ from the $matchList$. We also delete every block from every $matchList$ of every edge block that overlaps with $B_m$. Formally we delete a block B if
    $$
        B_m \cap B \neq [] \ \ \forall {B_i \in E, B \in matchList(B_i)}.
    $$
     We add $[0,v_i,v_k]$ to the $partitionList$ and $B_m$ to the $mappedList$.

    \textbf{\emph{step 3}}: If $(v_k + 1 ) \ MOD\ length(X) = v_s$ and the $mappedList$ obeys Equation \ref{eq2}, then we have found a common partition of $X$ and $Y$. The size of the partition is the length of the $partitionList$. Otherwise, we jump to the \emph{step 1}.

The construction is shown in Algorithm \ref{construction}.

\begin{algorithm}
\caption{constructSolution(i,$G_{cs}$)}
\label{construction}
\begin{algorithmic}
    \State $blockList$ = empty list of blocks
    \State $mappedList$ = empty list of blocks
    \State startpos = $\lfloor n/m \rfloor*i$
    \State k = startpos
    \Repeat
        \State addHeuristics($G_{cs}$,a,b)
        \State constructPDF(k,$G_{cs}$) using Equation \ref{pdf}
        \State B = choose an edge block from PDF
        \State M = choose a match block from $matchList(B)$
        \Comment Intelligent Positioning
        \State Update $matchList(B)$
        \State add B to $blockList$
        \State add M to the $mappedList$
        \State k = B.j + 1
    \Until{k $\neq$ startpos}
\end{algorithmic}
\end{algorithm}


\subsection{Intelligent Positioning }
    For every edge block of $G_{cs}$ in $X$, we have a $matchList$ that contains the matched block of string $Y$. In construction (step 1), when an edge block is chosen by the probability distribution, we take a block from the $matchList$ of the chosen edge block. We can choose the matched block randomly. But we observe that random choosing may lead to a very bad partition. For example, if ($X,Y$) = \{``ababc'',``abcab''\} then the $matchList([0,0,1]) = \{[1,0,1],[1,3,4]\}$. If we choose the first match block then eventually we will get the partition as \{``ab'',``ab'',``c''\} but a smaller partition exists and that is \{``ab'',``abc''\}.

    To overcome this problem, we have imposed a rule for choosing the matched block. We will select a block from the $matchList$ having the lowest possible span. Formally, for the  edge block, $B_i$, a block $B'\in matchList(B_i)$ will be selected such that $span(B')$ is the minimum.

    In our example $span([1,0,1]) = 3$ where as $span([1,3,4]) = 2$. So it is better to select the second block so that we do not miss the opportunity to match a larger block.
\subsection{Pheromone Update}
When each of the  ants in the colony construct a solution (i.e., a common partition), an iteration completes. We set the local best solution as the best partition that is the minimum length partition in an iteration. The global best solution for $n$ iterations is defined as the minimum length common partition over all the  $n$ iteration.

We define the fitness $F(L)$ of a solution $L$ as the reciprocal of the length of $L$. The pheromone of each interval of each target string is computed according to Equation \ref{pheromone} after each iteration. The pheromone values are bounded within the range $\tau_{MIN}$ and $\tau_{MAX}$. We update the pheromone values according to $L_{LB}$ or $L_{GB}$. Initially for the first 50 iterations we update pheromone by only $L_{LB}$ to favor the search exploration. After that we develop a scheduling where the frequency of updating with $L_{LB}$ decreases and $L_{GB}$ increases to facilitate exploitation. The pheromone update algorithm is listed in Algorithm \ref{update_pheromone_mcsp}

\begin{algorithm}
\caption{decreasePheromone(Blocklist E))}
\label{decrease}
\begin{algorithmic}
\ForAll{ Block B in E}
    \State pheromone(B) $\gets$ pheromone(B) - $\epsilon$ $\cdot$ pheromone(B)
\EndFor
\end{algorithmic}
\end{algorithm}

\begin{algorithm}
\caption{increasePheromone(Blocklist E))}
\label{increase}
\begin{algorithmic}
\ForAll{ Block B in E}
    \State pheromone(B) $\gets$ pheromone(B) + $\epsilon$ $\cdot$ $\frac{1}{|E|}$
\EndFor
\end{algorithmic}
\end{algorithm}

\begin{algorithm}
\caption{updatePheromoneSchedule(iterationCounter,$G_{cs}$,$L_{LB}$,$L_{GB}$)}
\label{update_pheromone_mcsp}
\begin{algorithmic}
    \State E $\gets$ edge blocks of $G_{cs}$
    \State decreasePheromone(E)

    \If{$iterationCounter \leq 50$}
        \State increasePheromone($L_{LB}$)
    \ElsIf{$iterationCounter \leq 100$}
        \If{$iterationCounter MOD 5 == 0$}
            \State increasePheromone($L_{LB}$)
        \Else
            \State increasePheromone($L_{GB}$)
        \EndIf
    \ElsIf{$iterationCounter \leq 200$}
        \If{$iterationCounter MOD 4 == 0$}
            \State increasePheromone($L_{LB}$)
        \Else
            \State increasePheromone($L_{GB}$)
        \EndIf
    \ElsIf{$iterationCounter \leq 400$}
        \If{$iterationCounter MOD 3 == 0$}
            \State increasePheromone($L_{LB}$)
        \Else
            \State increasePheromone($L_{GB}$)
        \EndIf
    \ElsIf{$iterationCounter \leq 800$}
        \If{$iterationCounter MOD 2 == 0$}
            \State increasePheromone($L_{LB}$)
        \Else
            \State increasePheromone($L_{GB}$)
        \EndIf
    \Else
          \State increasePheromone($L_{LB}$)

    \EndIf

    \State Update $tau_{max}$ and $tau_{min}$
    \ForAll{ Block B in E}
        \State Bound pheromone(B) between $tau_{max}$ and $tau_{min}$
    \EndFor

\end{algorithmic}
\end{algorithm}

\subsection{The Pseudocode}
The pseudocode of our approach for solving MCSP is given in Algorithm \ref{aco_mcsp}.
\begin{algorithm}
\caption{MMAS(X,Y)}
\label{aco_mcsp}
\begin{algorithmic}
\State $G_{cs}$ $\gets$ construct common substring graph of string X and Y
\For{$run =1 \to nRun$} \Comment $nRun$ $\gets$ number of Runs
    \State initialize($G_{cs}$)
    \State interationCounter = 0
    \Repeat
        \State iterationCounter = iterationCounter + 1;
        \State Initialize local best
        \For{$i =1 \to nAnts$}
            \State constructSolution(i,$G_{cs}$)
            \State update localBest ($L_{LB}$)
        \EndFor
        \State update globalBest ($L_{GB}$)
        \State updatePheromoneSchedule(iterationCounter,$G_{cs}$)
    \Until{time reaches $maxAllowedTime$ or No update found for $maxAllowedIteration$}
\EndFor
\end{algorithmic}
\end{algorithm}

\section{Experiments}
We have conducted our experiments in a computer with Intel Core 2 Quad CPU 2.33 GHz. The available RAM was 4.00 GB. The operating system was Windows 7. The programming environment was java. jre version is``1.7.0\_15''. We have used JCreator as the Integrated Development Environment. The maximum allowed time for test case instance was 120 minutes.

\subsection{Datasets}
We have conducted our experiments on two types of data: randomly generated DNA sequences and real gene sequences.
\subsubsection{Random DNA sequences:}
We have generated $30$ random DNA sequences each of length at most 600 using \cite{seq}. The fraction of bases $A$, $T$, $G$ and $C$ is assumed to be 0.25 each. For each DNA sequence we shuffle it to create a new DNA sequence. The shuffling is done using the online toolbox \cite{shuffle}. The original random DNA sequence and its shuffled pair constitute a single input ($X,Y$) in our experiment. This dataset is divided into 3 classes. The first 10 have lengths within [100-200] bps (base-pairs), the next 10 have lengths within $[201,400]$ and the rest 10 have lengths within $[401,600]$ bps.

\subsubsection{Real Gene Sequences:}
We have collected the real gene sequence data from the NCBI GenBank\footnote{http://www.ncbi.nlm.nih.gov}. For simulation, we have chosen Bacterial Sequencing (part 14). We have taken the first 15 gene sequences whose lengths are within $[200,600]$.
%

\subsection{Parameter Tuning}

There are several parameters which have to be carefully set to obtain good results. To obtain a good set of parameters we have done a preliminary experiment. In our experiment we have chosen 3 values for each of the parameters. so there are 243 possible permutations of the 5 parameters. The values of the parameters used in our experiment is listed in Table \ref{table:param1}. We have chosen 2 input cases from each of the groups (group1, group2, group3 and realgene). The time limits are set to 10, 20, 30 and 20 minutes for the 4 groups, respectively. The algorithm is run for 4 times and the average result is recorded. Let the partition size of each of the case is denoted by $A^{i}$ where $i \in [1,8]$. With these settings, we find rank of a permutation by the following rule:
$$
R_j = \sum_{i \in [1,8]} {A^{i}_{j}/max(A^{i})}\ \ \forall j \in [1,243]
$$

After computing the Rank, $R$, we find the permutation of the parameters for which the rank is minimum. The best found parameters are reported in Table \ref{table:param}. 

\begin{table}
\centering
\caption{List of Parameters. The first column represents the name, the second column represents the symbol of the parameter and the third column represent the set of values used for tuning}
\label{table:param1}
\begin{tabular}{|c|c|c|}
    \hline
    \textbf{Name} & \textbf{Symbol} & \textbf{value set}\\
    \hline
     Pheromone information & $\alpha$ & \{1,2,3\}\\
     Heuristic information & $\beta$ & \{3,5,10\}\\
    Evaporation rate & $\varepsilon$ & \{0.02,0.04,.05\}\\
     Number of Ants& $nAnts$ & \{20,60,100\}\\
     Probability of best solution & $p_{best}$ & \{0.005,0.05,0.5\}\\
    \hline
\end{tabular}
\end{table}

\begin{table}
\centering
\caption{Best found values of the parameters. The first column is the symbol of the parameter and the second column is the best found value}
\label{table:param}
\begin{tabular}{|c|c|}
    \hline
    \textbf{Parameters} & \textbf{Value}\\
    \hline
    $\alpha$ & $2.0$\\
    $\beta$ & $10.0$\\
    Evaporation rate, $\varepsilon$ & $0.05$\\
    $nAnts$ & 100\\
    $p_{best}$ & $0.05$\\
    $initPheromone$ & $10.0$\\
    Maximum Allowed Time & $120$ min\\
    \hline
\end{tabular}
\end{table}

\subsection{Results and Analysis}
We have compared our approach with the greedy algorithm of \cite{chrobak} because none of the other algorithms in the literature are for general MCSP: each of the other approximation algorithms put some restrictions on the parameters. As it is expected the greedy algorithm runs very fast. All of the result by greedy algorithm presented in this paper outputs within 2 minutes.

\subsubsection{Random DNA sequence:}
Table \ref{table:res_rand1}, Table \ref{table:res_rand2} and Table \ref{table:res_rand3} present the comparison between our approach and the greedy approach \cite{chrobak} for the random DNA sequences. For a particular DNA sequence, the experiment was run 15 times and the average result is reported. The first column under any group reports the partition size computed by the greedy approach, the second column is the average partition size found by MMAS, the third and fourth column report the worst and best results among 15 runs, the fifth column represents the difference between the two approaches. A positive (negative) difference indicates that the greedy result is better (worse) than the MMAS result by that amount. The sixth column reports the standard deviation of 15 runs of MMAS, the seventh column is the average time in second by which the reported partition size is achieved. The first 3 columns summarize the t-statistic result for greedy vs. MMAS. The first column reports the t-value of two sample t-test. A positive t-value indicate significant improvement. The second column presents the p-value. A lower p-value represent higher significant improvement and the third column reports whether the null hypothesis is rejected or accepted. Here the null hypothesis is that the two random population (partition sizes from greedy and MMAS) have equal means. We have used $+,-,\approx$ to denote improvement, deteriotion and almost equal respectively. According to t-statistic value with 5\% significance value we have found better solution in 28 cases for MMAS. For the other 2 case we got worse result in 5\% significance level.
\begin{sidewaystable}
\begin{center}
\caption {Comparison between Greedy approach \cite{chrobak} and MMAS on random DNA sequences (Group 1, [100-200] bps). Here, Difference = MMAS(Avg.) - Greedy. Best and Worst report the maximum and minimum partition size among 15 runs using MMAS. }
\label{table:res_rand1}
\scalebox{0.8}{
\begin{tabular}{ |l|l|l|l|l|l|l|l|l|l| }
  \hline
  \textbf{Greedy} & \textbf{MMAS(Avg.)} & \textbf{Worst} & \textbf{Best} & \textbf{Difference} & \textbf{Std.Dev.(MMAS} & \textbf{Time in sec(MMAS)} & \textbf{tstat} & \textbf{p-value} & \textbf{significance}\\
  \hline
    46 & 42.8667 & 43 & 42 & -3.1333 & 0.3519 & 114.6243 & 34.4886 & 0.0000 &  + \\
56 & 51.8667 & 52 & 51 & -4.1333 & 0.5164 & 100.823  & 31 & 0.0000 &  +\\
62 & 57 & 58 & 55 & -5 & 0.6547 & 207.5253  & 29.5804 & 0.0000 &  +\\
46 & 43.3333 & 43 & 43 & -2.6667 & 0.488 & 168.3098  & 21.166 & 0.0000 &  + \\
44 & 42.9333 & 43 & 43 & -1.0667 & 0.2582 & 42.7058  & 16 & 0.0000 &  +\\
48 & 42.8 & 43 & 42 & -5.2 & 0.414 & 75.2033  & 48.6415 & 0.0000 &  +\\
65 & 60.6 & 60 & 60 & -4.4 & 0.5071 & 131.9478  & 33.6056 & 0.0000 &  +\\
51 & 46.9333 & 47 & 47 & -4.0667 & 0.4577 & 201.2292  & 34.4086 & 0.0000 &  + \\
46 & 45.5333 & 46 & 45 & -0.4667 & 0.5164 & 172.6809   & 3.5 & 0.0016 &  + \\
63 & 59.7333 & 60 & 59 & -3.2667 & 0.7037 & 288.4226  & 17.9781 & 0.0000 &  +\\
  \hline

\end{tabular}
}
\end{center}
\end{sidewaystable}

\begin{sidewaystable}
\begin{center}
\caption {Comparison between Greedy approach \cite{chrobak} and MAX-MIN on random DNA sequences (Group 2, [201-400] bps). Here, Difference = MMAS(Avg.) - Greedy. Best and Worst report the maximum and minimum partition size among 15 runs using MMAS}
\label{table:res_rand2}
\scalebox{0.8}{
\begin{tabular}{ |l|l|l|l|l|l|l|l|l|l| }
  \hline
  \textbf{Greedy} & \textbf{MMAS} & \textbf{Worst} & \textbf{Best} & \textbf{Difference} & \textbf{Std.Dev.(MMAS)} & \textbf{Time in sec(MMAS)} & \textbf{tstat} & \textbf{p-value} & \textbf{significance}\\
  \hline
 119 & 113.9333 & 116 & 111 & -5.0667 & 1.3345 & 1534.1015 & 14.7042 & 0.0000 &  + \\
122 & 118.9333 & 121 & 117 & -3.0667 & 0.9612 & 1683.1146 & 12.3572 & 0.0000 &  +  \\
114 & 112.5333 & 114 & 111 & -1.4667 & 0.8338 & 1398.5315 & 6.8126 & 0.0000 &  + \\
116 & 116.4 & 117 & 115 & 0.4 & 0.7368 & 1739.3478 & -2.1026 & 0.0446 &  - \\
135 & 132.2 & 135 & 130 & -2.8 & 1.3202 & 1814.7264 & 8.2143 & 0.0000 &  + \\
108 & 106.0667 & 107 & 105 & -1.9333 & 0.8837 & 1480.2378 & 8.4731 & 0.0000 &  + \\
108 & 98.4 & 101 & 96 & -9.6 & 1.2421 & 1295.2485 & 29.9333 & 0.0000 &  +  \\
123 & 118.4 & 120 & 117 & -4.6 & 0.7368 & 1125.2353 & 24.1802 & 0.0000 &  + \\
124 & 119.4667 & 121 & 117 & -4.5333 & 1.0601 & 1044.4141 & 16.5622 & 0.0000 &  + \\
105 & 101.8667 & 103 & 101 & -3.1333 & 0.7432 & 1360.1529 & 16.328 & 0.0000 &  + \\
  \hline

\end{tabular}
}
\end{center}
\end{sidewaystable}

\begin{sidewaystable}
\begin{center}
\caption {Comparison between Greedy approach \cite{chrobak} and MAX-MIN on random DNA sequences (Group 3, [401-600] bps). Here, Difference = MMAS(Avg.) - Greedy. Best and Worst report the maximum and minimum partition size among 15 runs using MMAS}
\label{table:res_rand3}
\scalebox{0.8}{
\begin{tabular}{ |l|l|l|l|l|l|l|l|l|l| }
  \hline
  \textbf{Greedy} & \textbf{MMAS} & \textbf{Worst} & \textbf{Best} & \textbf{Difference} & \textbf{Std.Dev.(MMAS)} & \textbf{Time in sec(MMAS)} & \textbf{tstat} & \textbf{p-value} & \textbf{significance}\\
  \hline
    182	& 179.9333 & 181 & 177 & -2.0667 & 1.7099 & 1773.0398 & 4.6810 & 0.0001 & + \\
175 & 176.2000 & 177 & 175 & 1.2000 & 0.8619 & 3966.8293 & -5.3923 & 0.0000 & - \\
196 & 187.8667 & 189 & 187 & -8.1333 & 0.7432 & 1589.2953 & 42.3833 & 0.0000 & + \\
192 & 184.2667 & 185 & 184 & -7.7333 & 0.4577 & 2431.1580 & 65.4328 & 0.0000 & + \\
176 & 171.5333 & 173 & 171 & -4.4667 & 0.9155 & 1224.8943 & 18.8965 & 0.0000 & + \\
170 & 163.4667 & 165 & 160 & -6.5333 & 1.8465 & 1826.1438 & 13.7036 & 0.0000 & + \\
173 & 168.4667 & 170 & 167 & -4.5333 & 1.1872 & 1802.1655 & 14.7886 & 0.0000 & + \\
185 & 176.3333 & 177 & 175 & -8.6667 & 0.8165 & 1838.5603 & 41.1096 & 0.0000 & + \\
174 & 172.8000 & 175 & 172 & -1.2000 & 1.5675 & 4897.4688 & 2.9649 & 0.0061 & + \\
171 & 167.2000 & 168 & 167 & -3.8000 & 0.5606 & 1886.2098 & 26.2523 & 0.0000 & + \\
  \hline

\end{tabular}
}
\end{center}
\end{sidewaystable}

\subsubsection{Effects of Dynamic Heuristics:}
\begin{figure}
   \begin{center}
  \includegraphics[width=1\textwidth]{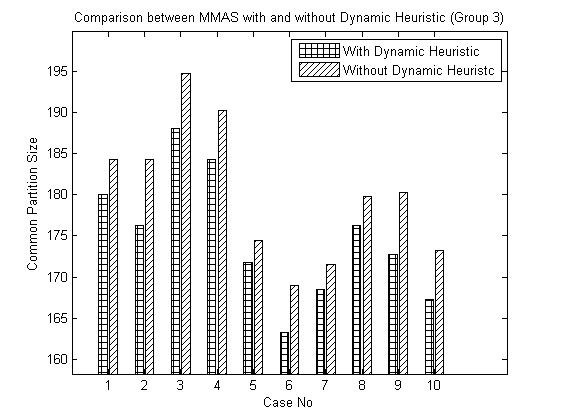}
  \caption{Comparison between MMAS with and without dynamic heuristic (Group 1)}
  \label{output200_dh}
  \end{center}
\end{figure}

\begin{figure}
  \begin{center}
  \includegraphics[width=1\textwidth]{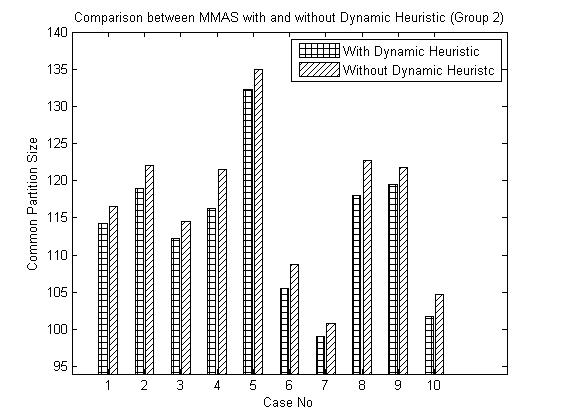}
  \caption{Comparison between MMAS with and without dynamic heuristic (Group 2)}
  \label{output400_dh}
  \end{center}
\end{figure}

\begin{figure}
  \begin{center}
  \includegraphics[width=1\textwidth]{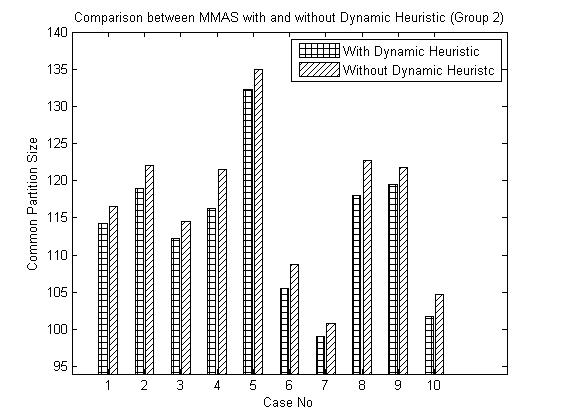}
  \caption{Comparison between MMAS with and without dynamic heuristic (Group 3)}
  \label{output600_dh}
  \end{center}
\end{figure}

In Section~\ref{sec:dynamicH}, we discussed the dynamic heuristic we employ in our algorithm. We conducted experiments to check and verify the effect of this dynamic heuristic. We conducted experiments with two versions of our algorithm- with and without applying the dynamic heuristic. The effect is presented in Table \ref{table:res_rand_dh}, where for each group the average partition size with dynamic heuristic and without dynamic heuristic is reported. The positive difference depicts the improvement using dynamic heuristic. Out of 30 cases we found positive differences on 27 cases. This clearly shows the significant improvement using dynamic heuristics. It can also be observed that with the increase in length, the positive differences are increased. Figures \ref{output200_dh}, \ref{output400_dh}, and \ref{output600_dh} show the case by case results. The blue bars represent the partition size using dynamic heuristic and the red bars represent the partition size without the dynamic heuristic.

\begin{table}
\begin{center}
\caption {Comparison between MMAS with and without dynamic heuristic on random dna sequence}
\label{table:res_rand_dh}
\scalebox{0.6}{
\begin{tabular}{ |l|l|l|l|l|l|l|l|l| }
  \hline
  \multicolumn{3}{|c|}{Group 1 (200 bps)} & \multicolumn{3}{|c|}{Group 2 (400 bps)} & \multicolumn{3}{|c|}{Group 3 (600 bps)}\\
  \hline
  \textbf{MMAS} & \textbf{MMAS(w/o heuristic)} & \textbf{Difference} & \textbf{MMAS} & \textbf{MMAS(w/o heuristic)} & \textbf{Difference} & \textbf{MMAS} & \textbf{MMAS(w/o heuristic)} & \textbf{Difference} \\
  \hline
 42.7500 & 43.2500 & 0.5000 & 114.2500 & 115.5000 & 1.2500 & 180.0000 & 183.2500 & 3.2500 \\
51.5000 & 50.7500 & -0.7500 & 119.0000 & 121.0000 & 2.0000 & 176.2500 & 183.2500 & 7.0000 \\
56.7500 & 56.5000 & -0.2500 & 112.2500 & 113.5000 & 1.2500 & 188.0000 & 193.7500 & 5.7500 \\
43.0000 & 44.0000 & 1.0000 & 116.2500 & 120.5000 & 4.2500 & 184.2500 & 189.2500 & 5.0000 \\
43.0000 & 42.7500 & -0.2500 & 132.2500 & 134.0000 & 1.7500 & 171.7500 & 173.5000 & 1.7500 \\
42.2500 & 42.5000 & 0.2500 & 105.5000 & 107.7500 & 2.2500 & 163.2500 & 168.0000 & 4.7500 \\
60.0000 & 60.5000 & 0.5000 & 99.0000 & 99.7500 & 0.7500 & 168.5000 & 170.5000 & 2.0000 \\
47.0000 & 47.5000 & 0.5000 & 118.0000 & 121.7500 & 3.7500 & 176.2500 & 178.7500 & 2.5000 \\
45.7500 & 46.0000 & 0.2500 & 119.5000 & 120.7500 & 1.2500 & 172.7500 & 179.2500 & 6.5000 \\
59.2500 & 61.5000 & 2.2500 & 101.7500 & 103.7500 & 2.0000 & 167.2500 & 172.2500 & 5.0000 \\
  \hline

\end{tabular}
}
\end{center}
\end{table}

\subsubsection{Real Gene Sequence:}
Table \ref{table:real} shows the minimum common partition size found by our approach and the greedy approach for the real gene sequences. Out of 15 cases positive improvement is found in 10 cases in 5\% significance level.


\begin{sidewaystable}
\begin{center}
\caption{Comparison between Greedy approach \cite{chrobak} and MMAS on real gene sequence.Here, Difference = MMAS(Avg.) - Greedy. Best and Worst report the maximum and minimum partition size among 15 runs using MMAS}
\label{table:real}
\scalebox{0.8}{
\begin{tabular}{|l|l|l|l|l|l|l|l|l|l|}
\hline
\textbf{Greedy} & \textbf{MMAS} & \textbf{Worst} & \textbf{Best} & \textbf{Difference} & \textbf{Std.Dev(MMAS)} & \textbf{Time in sec(MMAS)} & \textbf{tstat} & \textbf{p-value} & \textbf{significance}\\
\hline
95 & 87.66666667 & 88 & 87 & -7.333333333 & 0.487950036 & 863.8083333 & 58.2065 & 0.0000 & +\\
161 & 156.3333333 & 162 & 154 & -4.666666667 & 2.350278606 & 1748.34 & 7.6901 & 0.0000 & +\\
121 & 117.0666667 & 118 & 116 & -3.933333333 & 0.883715102 & 1823.4922 & 17.2383 & 0.0000 & +\\
173 & 164.8666667 & 167 & 163 & -8.133333333 & 1.187233679 & 1823.012533 & 26.5325 & 0.0000 & +\\
172 & 170.3333 & 172 & 169 & 1.2 & 1.207121724 & 2210.153533 & 3.8501 & 0.0006 & +\\
153 & 146 & 148 & 143 & -7 & 1.309307341 & 1953.838267 & 20.7063 & 0.0000 & +\\
140 & 141 & 142 & 140 & 1 & 0.755928946 & 2439.0346 & -5.1235 & 0.0000 & -\\
134 & 133.1333333 & 136 & 130 & -0.866666667 & 1.807392228 & 1406.804533 & 1.8571 & 0.0738 & $\approx$\\
149 & 147.5333333 & 150 & 145 & -1.466666667 & 1.505545305 & 2547.519267 & 3.7730 & 0.0008 & +\\
151 & 150.5333333 & 152 & 148 & -0.466666667 & 1.597617273 & 1619.6364 & 1.1313 & 0.2675 & $\approx$\\
126 & 125 & 127 & 123 & -1 & 1 & 1873.3868 & 3.8730 & 0.0006 & +\\
143 & 139.1333333 & 141 & 137 & -3.866666667 & 1.245945806 & 2473.249067 & 12.0194 & 0.0000 & +\\
180 & 181.5333333 & 184 & 179 & 1.533333333 & 1.35576371 & 2931.665333 & -4.3802 & 0.0002 & -\\
152 & 149.3333333 & 151 & 147 & -2.666666667 & 1.290994449 & 2224.403733 & 8.0000 & 0.0000 & +\\
157 & 161.6 & 164 & 160 & 4.6 & 1.242118007 & 1739.612133 & 1-14.3430 & 0.0000 & -\\
  \hline
\end{tabular}
}
\end{center}
\end{sidewaystable}





\section{Conclusion}
Minimum Common String Partition problem has important applications  in computational biology. In this paper, we have described a metaheuristic approach to solve the problem. We have used static and dynamic heuristic information in this approach with intelligent positioning. The simulation is conducted on random DNA sequences and real gene sequences. The results are significantly better than the previous results. The t-test result also shows significant improvement. As a future work different other metaheuristic techniques may be applied to present better solutions to the problem.

\bibliographystyle{splncs}
\bibliography{Bibliography}

\end{document}